\documentclass{article}
\usepackage{float}
\usepackage{spconf,amsmath,graphicx}

\usepackage{booktabs}
\usepackage{multirow}
\usepackage{tabularx}

\usepackage{anyfontsize}
\newlength\stdbls % standard baselineskip
\AtBeginDocument{%
  \setlength\stdbls{\baselineskip}%
  \gdef\stdblsn{\strip@pt\stdbls}%
} 

\usepackage{arydshln}
\usepackage[colorlinks=true, allcolors=blue]{hyperref}

\usepackage[dvipsnames]{xcolor}

\usepackage{tikz}
\usetikzlibrary{positioning, patterns}

\definecolor{cyan}{HTML}{88CCEE}
\definecolor{teal}{HTML}{44AA99}
\definecolor{sand}{HTML}{DDCC77}
\definecolor{wine}{HTML}{882255}
\definecolor{purple}{HTML}{AA4499}
\definecolor{pale_gray}{HTML}{DDDDDD}

\newcommand{\figurewidth}[0]{6}
\tikzset{
    fixed_width/.style={rectangle, very thick, minimum height=0.55cm, minimum width=\figurewidth cm},
    loss/.style={color=darkgray!80, fixed_width},
    model_node/.style={draw=cyan, minimum height=1cm, pattern=north east lines, pattern color=cyan!10, fixed_width},
    head_node/.style={draw=teal!80, fill=teal!5, fixed_width},
    table/.style={rectangle, draw=black!40, fill=black!5, text=black!60, pattern=dots, pattern color=lightgray!10, thick},
    table_node/.style={table, minimum size=0.5cm},
    io_node/.style={rounded corners, fixed_width, table},
    invisible_node/.style={rectangle, minimum height=0.55cm, text=black!80, minimum width=\figurewidth cm}
}

\usepackage{todonotes}

\def\x{{\mathbf x}}

\title{On the Use of Semantically-Aligned Speech Representations \\
for Spoken Language Understanding}
\name{Gaëlle Laperrière$^1$, Valentin Pelloin$^2$, Mickaël Rouvier$^1$, Themos Stafylakis$^3$, Yannick Estève$^1$}

\address{ $^1$ LIA - Avignon Université, France\\
 $^2$ LIUM - Le Mans Université, France \\
 $^3$ Omilia - Conversational Intelligence, Greece}

\begin{document}
%\ninept
%
\maketitle
\begin{abstract}
In this paper we examine the use of semantically-aligned speech representations for end-to-end spoken language understanding (SLU).
We employ the recently-introduced SAMU-XLSR model, which is designed to generate a single embedding that captures the semantics at the utterance level, semantically aligned across different languages. This model combines the acoustic frame-level speech representation learning model (XLS-R) with the Language Agnostic BERT Sentence Embedding (LaBSE) model.
We show that the use of the SAMU-XLSR model instead of the initial XLS-R model improves significantly the performance in the framework of end-to-end SLU.
Finally, we present the benefits of using this model towards language portability in SLU.

\end{abstract}
\section{Introduction}
\label{sec:intro}

Spoken language understanding (SLU) refers to natural language processing tasks related to semantic extraction from speech~\cite{tur2011spoken}.
Different tasks can be addressed as SLU tasks, such as named entity recognition from speech, call routing, slot filling task in a context of human-machine dialogue. 

To our knowledge, end-to-end neural approaches have been proposed four years ago in order to directly extract the semantics from speech signal, by using a single neural model~\cite{ghannay2018end, haghani2018audio,serdyuk2018towards}, instead of applying a classical cascade approach based on the use of an automatic speech recognition (ASR) system, followed by a natural language understanding processing (NLU) module applied to the automatic transcription~\cite{tur2011spoken}.
Two are the main advantages of end-to-end approaches. The first one is related to the joint optimization of the ASR and NLU part, since the unique neural model is optimized only for the final SLU task.
The second one is the mitigation of error propagation: when using a cascade approach, errors generated by the first modules propagate to the following ones. 

Since 2018, end-to-end approaches have became very popular in the SLU literature~\cite{desot2019slu,dinarelli2020data,radfar2020end,palogiannidi2020end,poncelet2021low}.
A main issue of these approaches is the lack of bimodal annotated data (speech audio recordings with semantic manual annotation). Several methods have been proposed in order to address this issue, e.g. transfer learning techniques~\cite{bhosale2019end,Caubriere2019}, \cite{huang2020leveraging} or artificial augmentation of the training data using speech synthesis~\cite{desot2020corpus,lugosch2020using}.

Self-supervised learning (SSL), that benefits from unlabelled data, recently opened new perspectives for automatic speech recognition and natural language processing~\cite{baevski2020wav2vec,devlin-etal-2019-bert}. 
SSL has been successfully applied to several SLU tasks, especially through cascade approaches~\cite{laperriere2021we}: the ASR system benefits from learning better speech unit representations~\cite{liu2020mockingjay,liu2021tera,hsu2021hubert} while the NLU module benefits from BERT-like models~\cite{devlin-etal-2019-bert}.
The use of an end-to-end approach exploiting directly both speech and text SSL models is limited by the difficulty to unify the speech and textual representation spaces, in addition to the complexity of managing a huge number of model parameters.
Some approaches have been proposed in order to exploit the BERT-like capabilities within an end-to-end SLU model, e.g. by projecting some kinds of sequences of embeddings extracted by an ASR sub-module to a BERT model~\cite{wang2020large,chung2021splat}, or by tying at the sentence level the acoustic embeddings to a SLU fine-tuned BERT model for a speech intent detection task~\cite{huang2020leveraging,agrawal2022tie}.
In~\cite{muller2021pursuit}, a similar approach is extended in order to build a multilingual end-to-end SLU model, again for speech intent detection.

Earlier this year, a new promising model was introduced in~\cite{khurana2022samu}. 
The model combines a state-of-the-art multilingual acoustic frame-level speech representation learning model XLS-R~\cite{babu2021xls} with the Language Agnostic BERT Sentence Embedding~\cite{feng2022language} (LaBSE) model to create an utterance-level multimodal multilingual speech encoder. 
This model is named SAMU-XLSR, for Semantically-Aligned Multimodal Utterance-level Cross-Lingual Speech Representation learning framework.

In this paper, we analyze the performance and the behavior of the SAMU-XLSR model using the French MEDIA benchmark dataset, which is considered as a very challenging benchmarks for SLU~\cite{bechet2019benchmarking}.
Moreover, by using the Italian PortMEDIA corpus~\cite{lefevre2012leveraging}, we also investigate the potential of porting an existing end-to-end SLU model from one language (French) to another (Italian) through two scenarios concerning the target language: zero-shot or low-resource learning.

\section{SAMU-XLSR}
\label{sec:sameer_xlsr}

Self-supervised representation learning (SSL) approaches such as Wav2Vec-2.0~\cite{baevski2020wav2vec}, HuBERT~\cite{hsu2021hubert}, and WavLM~\cite{chen2022wavlm} aim to provide powerful deep feature learning (speech embedding) without requiring large annotated datasets. 
Speech embeddings are extracted at the acoustic frame-level i.e. for short speech segments of 20 ms duration, and they can be used as input features to a model that is specific for the downstream task. 
%These speech encoder models can also be fine-tuned by supervised learning on such downstream tasks, especially when the amount of labelled data is low.
These speech encoders have been successfully used in several tasks, such as automatic speech recognition~\cite{baevski2020wav2vec}, speaker verification~\cite{chen2022large,chen2022does} and emotion recognition~\cite{macary2021use,pepino2021emotion}. 
Self-supervision learning for such speech encoders is designed to discover speech representations that encode pseudo-phonetic or phonotactic information rather than high-level semantic information~\cite{sanabria2022measuring}.
On the other hand, high-level semantic information is particularly useful in some tasks such as Machine Translation (MT) or Spoken Language Understanding (SLU). In~\cite{khurana2022samu}, the authors propose to address this issue using a new framework called SAMU-XLSR, which learns semantically-aligned multimodal utterance-level cross-lingual speech representations.

\begin{figure}[htb!]
\centering

\includegraphics[width=0.9\columnwidth]{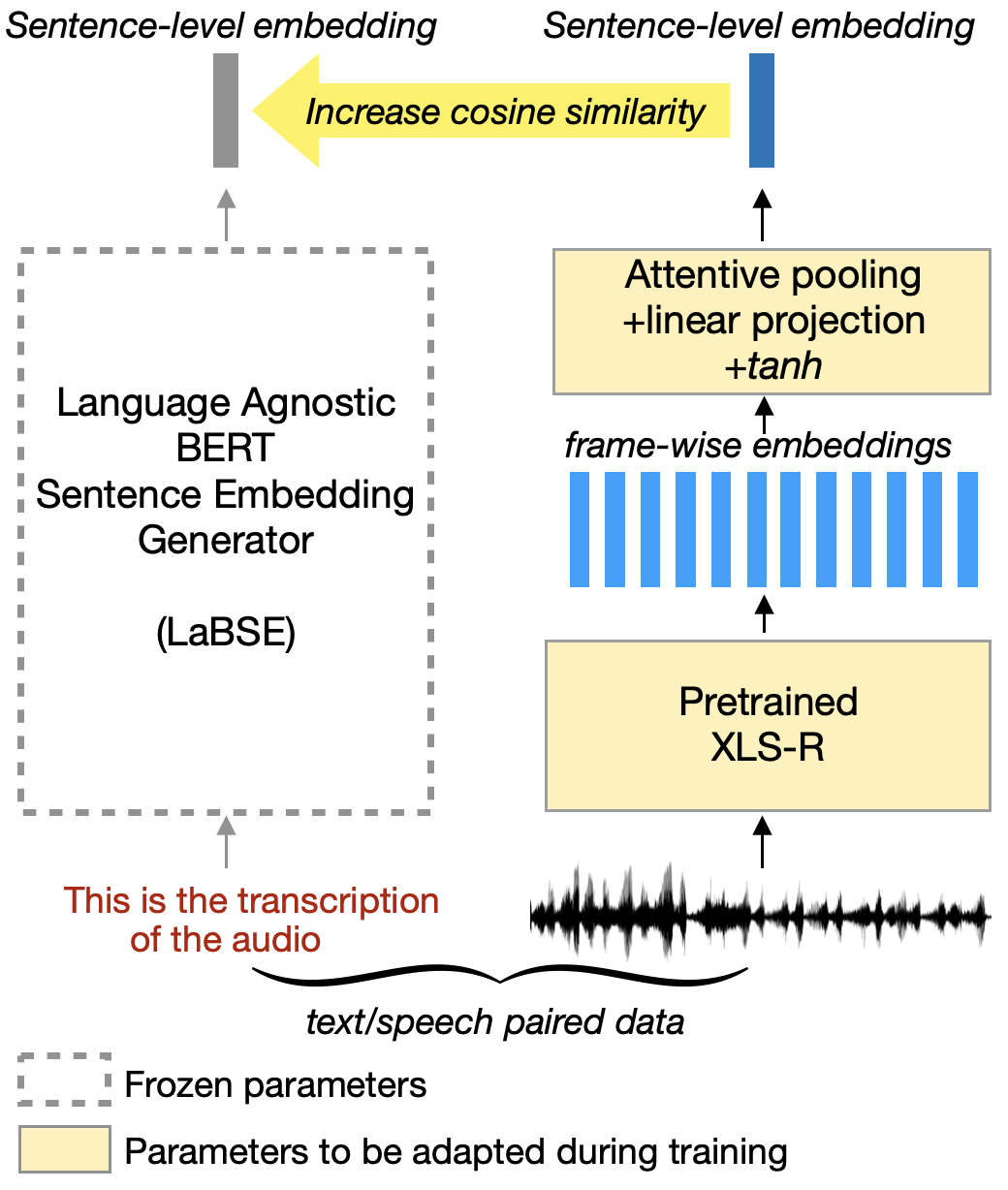}
\caption{Training process of SAMU-XLSR.} 
\label{fig:trainingSAMUXLSR}
%\vspace{-0.5cm}
\end{figure}

SAMU-XLSR is based on the pre-trained multilingual XLS-R~\footnote{https://huggingface.co/facebook/wav2vec2-xls-r-300m}~\cite{babu2021xls} on top of which all the embeddings generated by processing an audio file are connected to an attentive pooling module.

Thanks to this pooling mechanism (which is followed by linear projection layer and the \textsl{tanh} function), the frame-level contextual representations are transformed into a single utterance-level embedding vector. Figure~\ref{fig:trainingSAMUXLSR} summarizes the training process of the SAMU-XLSR model. 
Notice than the weights from the pre-trained XLS-R model continue being updated during the process.

The utterance-level embedding vector of SAMU-XLSR is trained via knowledge distillation from the pre-trained language agnostic LaBSE model~\cite{feng2022language}.
The LaBSE model\footnote{https://huggingface.co/sentence-transformers/LaBSE} has been trained on 109 languages and its text embedding space is semantically aligned across these 109 languages.
LaBSE attains state-of-the-art performance on various bi-text retrieval/mining tasks, while yielding promising zero-shot performance for languages not included in the training set (probably thanks to language similarities).  %and, probably thanks to language similarities, and is able to be successfully applied to these tasks for languages that do not have training data. 
Thus, given a spoken utterance, the parameters of SAMU-XLSR are trained to accurately predict a text embedding provided by the LaBSE text encoder of its corresponding transcript.
Because LaBSE embedding space is semantically-aligned across various languages, the text transcript would be clustered together with its text translations. 

By pulling the speech embedding towards the anchor embedding, cross-lingual speech-text alignments are automatically learned without ever seeing cross-lingual associations during training.
This property is particularly interesting in the SLU context in order to port an existing model built on a well-resourced language to another language with zero or low resources for training.

\section{Application to Spoken Language Understanding}

As defined in~\cite{tur2011spoken}, \textsl{spoken language understanding is the interpretation of signs conveyed by a speech signal}. 
This interpretation refers to a semantic representation manageable by computers.
Usually, this semantic representation is dedicated to an application domain that restricts the semantic field.
With the massive deployment of voice assistants like Apple's Siri, Amazon Alexa, Google Assistant, etc. a lot of recent papers aim to process speech intent detection as an SLU task~\cite{huang2020leveraging,desot2020corpus,lugosch2020using,muller2021pursuit,agrawal2022tie}.
In such a task, only one speech intent is generally expected by sentence: the speech intent detection task could be considered as a classification task at the sentence-level and, in addition, the SLU model has to fill some expected slots corresponding to the detected intent.

SLU benchmarks related to task-driven human-machine spoken dialogue can be more or less complex, depending on the richness of the semantic representation.
In this study, we focus on a hotel booking scenario through a telephone conversation, where the semantic representation is not related to speech intent detection, but based on a more complex ontology that derives from frames~\cite{baker1998berkeley}.
%SLU can be perceived as a machine translation task, that translates a spoken natural language to an artificial  language.

Our experimental work is carried out on the MEDIA SLU benchmark, described in section~\ref{subsec:MEDIA}.
We first expect to evaluate the performance of SAMU-XLSR used as a frame-wise feature extractor in comparison to the use of the initial XLS-R.
Then we analyse the quality of the semantic encoding for each layer of the SAMU-XLSR and XLS-R model, to better understand the impact of the SAMU-XLSR training on the XLS-R model.
We continue this investigation by fine-tuning the SAMU-XLSR and the XLS-R models on the downstream task.
We also investigate the capability of SAMU-XLSR to transfer the semantic knowledge captured on French data to Italian data related to the same SLU task, thanks to the PortMEDIA corpus described in section~\ref{subsec:PORTMEDIA}.
Last, we focus on the sentence-level embedding produced by the SAMU-XLSR model in order to measure the relevance of its semantic content to the target task, including in a language portability scenario and in a cross-modal setting.

\section{Data}

\subsection{The MEDIA benchmark}
\label{subsec:MEDIA}

The French MEDIA benchmark \cite{bonneau2009media} was created in 2002 as a part of a French governmental project named Technolangue. 
The \textsl{MEDIA Evaluation Package}\footnote{\url{http://catalog.elra.info/en-us/repository/browse/ELRA-E0024/}}
\footnote{International Standard Language Resource Number: 699-856-029-354-6} is distributed by ELRA and freely accessible for academic research.
Apart from the data itself, it defines a protocol for evaluating SLU modules, with a task of semantic extraction from speech in a context of human-machine dialogues.

The Wizard-of-Oz method has been used to create the dataset, consisting of hotel reservation phone call recordings. 
A human plays the role of a machine, by interacting with the user who believes that he is actually speaking to an intelligent machine. 
1258 official recorded dialogues were generated from around 250 speakers.
Only the user's turns are semantically annotated with both semantic annotation and transcription.
Table \ref{tab:media} presents the data distribution, in hours of speech and number of words, into the official training, development and test corpora.

\begin{table}[!ht]
    \begin{center}
        \begin{tabular}{| c | c | c | c |}
            \cline{2-4}    
            \multicolumn{1}{c |}{} & \textbf{Train} & \textbf{Dev} & \textbf{Test} \\
            \hline   
            \textbf{Hours} & 10h52m & 01h13m & 03h01m \\
            \hline   
            \textbf{Words} & 94.5k & 10.8k & 26.6k \\
            \hline  
        \end{tabular}
        \caption{Data distribution of the MEDIA corpus.}
        \label{tab:media}
    \end{center}
\end{table}

The semantic dictionary defined in MEDIA includes 83 basic attributes (renamed as \textsl{concepts} in our study) -- including 73 database attributes, 4 modifiers, and 6 general attributes -- and 19 specifiers~\cite{bonneau-maynard-etal-2006-results}:
\textit{room-number, hotel-name, location} are examples of database attributes, \textit{comparative, relative-distance} are examples of modifiers,  \textit{proposition-connector} and \textit{attribute-connector} are examples of general attributes, and \textit{address, travel} are examples of specifiers, that specializes the attribute role in a dialogue. 
Some complex linguistic phenomena, like co-references, are also managed thanks to this mechanism.
By combining attributes and specifiers, the total number of possible attribute/specifier pairs is 1121. 
In this study, the ``full" MEDIA version has been used for all experiments. Compared to the ``relax" one, the full's semantic annotations includes the use of specifiers and modifiers. 150 different semantic concepts (\textsl{i.e.} attribute/specifier pairs) are present in this version of MEDIA.

Each semantic concept is associated to values, also called word-support in the MEDIA documentation. 
The following translated sentence is an example of utterance present in the MEDIA dataset: ``I would like to book one double room in Paris". 
In this paper, we use annotations containing the transcript, the concepts and the location information of their values: ``I (would like to book, \textsl{reservation}), (one, \textsl{room-number}), (double room, \textsl{room-type}) in (Paris, \textsl{city})". 
%The concept ``city" will be linked to its value ``Paris".
%We can relate to the task as a contextual named entity recognition. 

%This benchmark aspires to a better evaluation of written and oral language technologies and the production and dissemination of language resources.

\subsection{The Italian PortMEDIA corpus}
\label{subsec:PORTMEDIA}
The Italian PortMEDIA corpus~\cite{lefevre2012leveraging} has been produced on the same hotel reservation task as MEDIA, and follows the same specifications. ELRA was in charge of the data collection and is currently distributing the corpus, as for MEDIA. \footnote{\url{http://www.elra.info/en/projects/archived-projects/port-media/}}

\begin{table}[!ht]
    \begin{center}
        \begin{tabular}{| c | c | c | c |}
            \cline{2-4}    
            \multicolumn{1}{c |}{} & \textbf{Train} & \textbf{Dev} & \textbf{Test} \\
            \hline 
            \textbf{Hours} & 07h18m & 02h32m & 04h51m \\
            \hline 
            \textbf{Words} & 21.7k & 7.7k & 14.7k \\
            \hline  
        \end{tabular}
        \caption{Data distribution of the PortMEDIA corpus.}
        \label{tab:portmedia}
    \end{center}
\end{table}

The Italian PortMEDIA corpus is made of 604 dialogues from more than 150 Italian speakers. 
Table \ref{tab:portmedia} gives information about the number of recorded hours and number of words distributed into training, development and test datasets. 
The PortMEDIA training corpus is more than four times smaller than the MEDIA one in terms of words.
If speech duration seems not so low in comparison, this is due to a less precise speech segmentation that includes large portions of silence. 

%\gl{And here} YE: not necessary on my opinion
The PortMEDIA corpus is only available with ``full" semantic annotations. 139 different concepts are present in the whole Italian PortMEDIA dataset.

%As the PortMEDIA name stands for Portability MEDIA, this corpus has been created for tasks of semantic knowledge transfer between French and Italian. It aims to permit SLU models to take benefit from these multilingual data, which are sharing the same semantic scheme. 
The PortMEDIA corpus is used in our study in order to conduct experiments on language portability from French to Italian for SLU. 

\subsection{MEDIA and PortMEDIA Metrics}
%
%SLU systems can be evaluated with different metrics. 
Historically, on the MEDIA corpus, two metrics are jointly used: the Concept Error Rate (CER) and the Concept-Value Error Rate (CVER). The CER is computed similarly to Word Error Rate (WER), by only taking into account the concepts occurrences in both the reference and the hypothesis files. The CVER metrics is an extension of the CER. It considers the correctness of the complete concept/value pair.

Since our models generate transcript with semantic concepts, we also evaluate our systems in terms of Character Error Rate (ChER) and WER. 
The ChER is computed by taking into account all the characters in the prediction that are not related to concepts (tags or concepts themselves). 
The same holds for WER but for words instead of characters.

For the semantic analysis of the sentence levels embeddings in section \ref{sec:boc-semantic-analysis}, we evaluate our bag-of-concepts outputs with the micro F$_1$-score, i.e. the harmonic mean between the precision and recall.

\section{Experiments}

\subsection{Layer-wise analysis of frame-level embeddings}

%[YE: why?] \cite{laperriere2022spoken} 
% GL : Idk, wasn't me 

%In this paper, we analyze how well is the semantic captured in the SAMU-XLSR’s signal embeddings, after forcing its outputs to be closer to LaBSE’s text embeddings.

\begin{figure}[ht!]
    \centering
    \begin{tikzpicture}
    \node[io_node] (inputs) {Input WAV};
    \node[model_node, above=0.15cm of inputs, align=center] (model) {Frozen or Fined-Tuned speech encoder\\(XLS-R or SAMU-XLSR)};

    \foreach \x in {0,...,11}{
        \node[table_node, draw=black!10, thick, minimum height=0.55cm, above right=0.3cm and (\x * 0.5cm) - \figurewidth cm -\pgflinewidth of model] (embd\x) {};
    }
    \node[invisible_node, text=darkgray!85, above=0.3cm of model] (seq_embedding) {Embeddings};

    \node[head_node, above=0.3cm of seq_embedding] (lstm) {bi-LSTM x3};
    \node[head_node, above=-\pgflinewidth of lstm] (fc) {Fully Connected x3};
    \node[head_node, above=-\pgflinewidth of fc] (softmax) {Softmax};

    \newcommand*\OutputsSeqList{{"$h$", "$h$", "$h$", "$e$", "$e$", "$l$", "$l$", "$\epsilon$", "$l$", "$l$", "$o$", "$o$"}}

    \foreach \x in {0,...,11}{
        \node[table_node, minimum height=0.55cm, above right=0.15cm and (\x * 0.5cm) - \figurewidth cm -\pgflinewidth of softmax] (output\x) {\scriptsize\strut\pgfmathprint{\OutputsSeqList[\x]}};
    }
    \node[invisible_node, above=0.15cm of softmax] (outputs) {};

    \node[loss, above=0cm of outputs] (outputs_name) {Output sequence};

    \draw [->] (inputs) edge (model);
    \draw [->] (model) edge (seq_embedding);
    \draw [->] (seq_embedding) edge (lstm);
    \draw [->] (softmax) edge (outputs);
    
    \end{tikzpicture}
    \caption{Neural Architecture for an SLU layer-wise analysis of speech encoders with the MEDIA dataset.}
    \label{fig:architecture}
\end{figure}
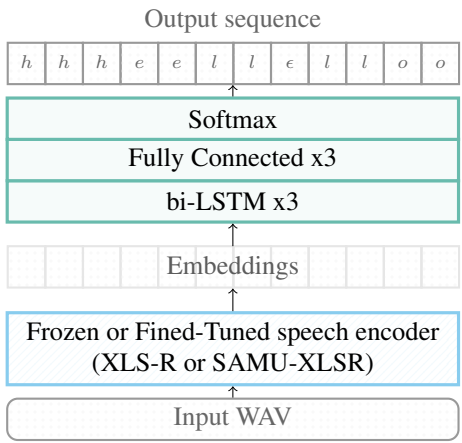

Figure~\ref{fig:architecture} presents the general architecture of the end-to-end model used for this study. 

This model aims to produce a transcription with semantic labels, such as: \textsl{I \textnormal{{\textless}reservation\textgreater} would like to book
\textnormal{{\textless}/reservation\textgreater} \textnormal{{\textless}room-number\textgreater}
one \textnormal{{\textless}/room-number\textgreater}
\textnormal{{\textless}room-type\textgreater} double room \textnormal{{\textless}/room-type\textgreater} in \textnormal{{\textless}city\textgreater} Paris \textnormal{{\textless}/city\textgreater}}.

For comparison purposes, we used as an encoder the original XLS-R or the SAMU-XLSR, taking WAV signal as input. 
Both have been frozen or fine-tuned during our experiments. 
The encoder is followed by three bi-LSTM layers of 1024 neurons, to keep the context of the entire segment when decoding the speech embedding. 
Then, the bi-LSTM outputs are fed to three linear layers of the same number of neurons, with a separate Adadelta optimizer starting with a $1.0$ learning rate.
When fine-tuning the encoder, the Adam optimizer is used, with a initial learning-rate of $0.0001$. The bi-LSTM layers have a similar optimizer.
All the decoding layers are activated with a LeakyReLU function.
The outputs are finally passing through a Softmax function and a CTC loss function is applied.
For early stopping and for the optimizers, the Concept Error Rate is the value we aim to minimize.

To make a layer-wise analysis, we removed the upper layers of each encoder, one by one, and extracted our speech embeddings.% from it. 
%We want to compare its embeddings extracted from each of its 24 layers to the ones extracted from the original XLS-R. 
%To do so, we removed all upper layers of each speech encoder, one by one. 
The encoder kept layers are frozen with their initial weights for the ``Frozen" architecture, or fine-tuned by supervision to solve the MEDIA task, leading to the ``Fine-Tuned" results. 
%The ChER evolution being similar to the WER one and the WER being a better visualization of what we aim for in an ASR task, we will skip the ChER analysis and focus on the WER.

\begin{figure}[htb!]
	\centering
	\includegraphics[width=\columnwidth]{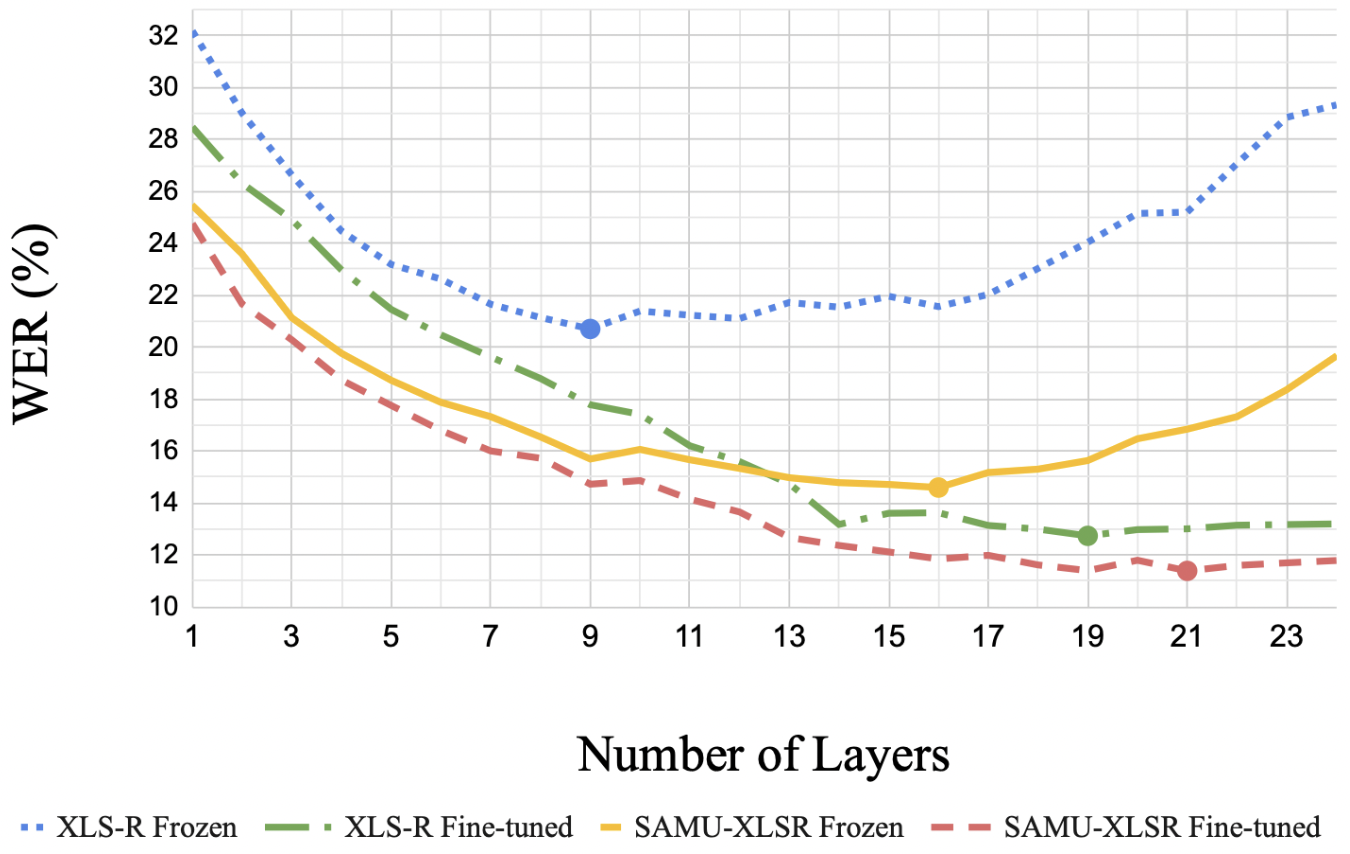}
	\caption{Layer-wise analysis of WER on the test data.}
	\label{fig:WER}
\end{figure}

Figure \ref{fig:WER} illustrates how the linguistic information is encoded through each layer of both encoders. 
First, we observe that in terms of WER, SAMU-XLSR gets better results than XLS-R. 
We also can see that the minimum WER is achieved with higher layers for SAMU-XLSR than it is for XLS-R, both frozen and fine-tuned. 
We assume this is due to the fine-tuning made on SAMU-XLSR by forcing its highest representations to be aggregated to LaBSE's text embeddings.

\begin{figure}[htb]
	\centering
	\includegraphics[width=\columnwidth]{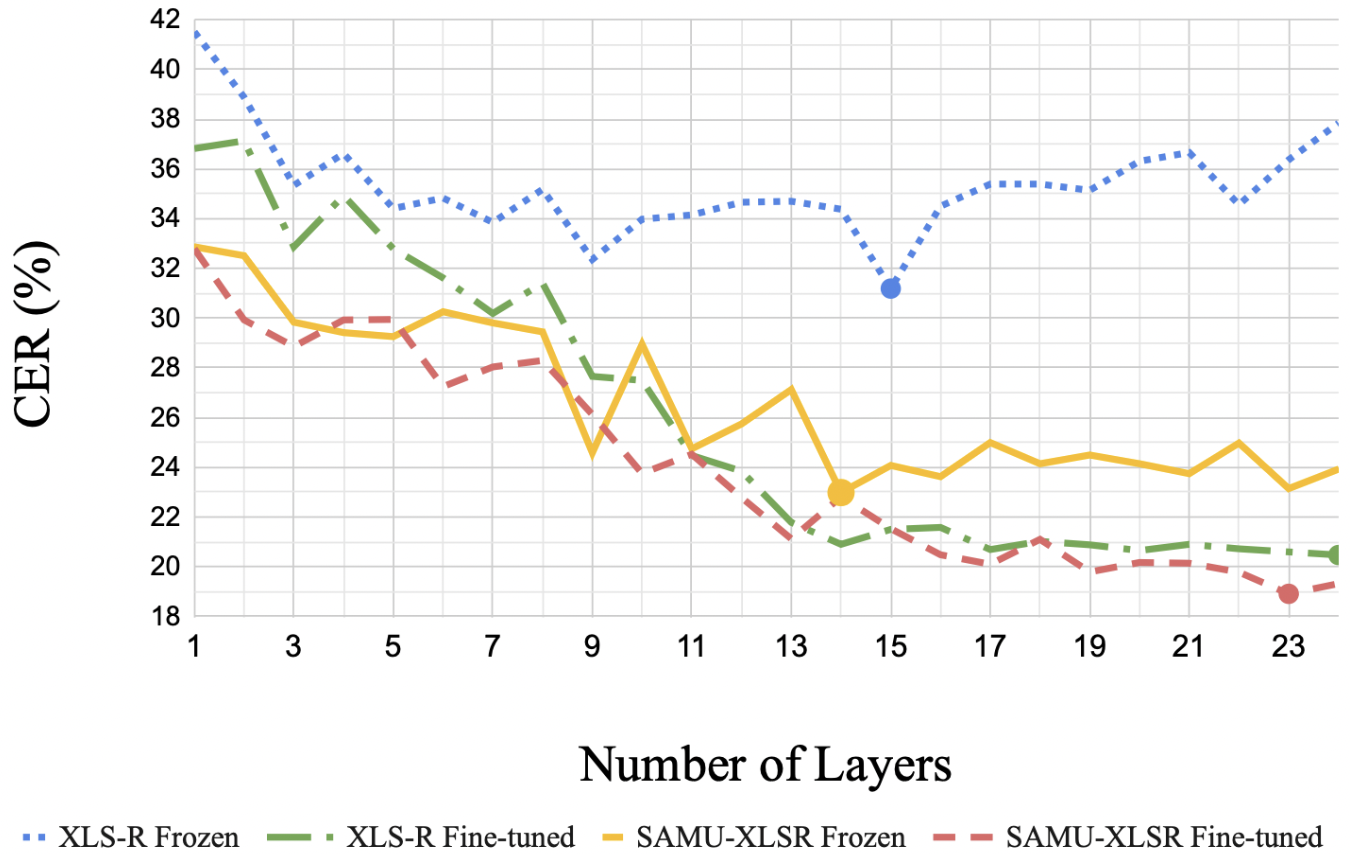}
	\caption{Layer-wise analysis of Concept Error Rate on the test data.}
	\label{fig:CER}
\end{figure}

Figure~\ref{fig:CER} presents results measured with the Concept Error Rate metric, relevant to the specific semantics of the downstream task. 

%\begin{figure}[htb]
%	\centering
%	\includegraphics[width=\columnwidth]{CVER4.png}
%	\caption{Layer-wise analysis of Concept Value Error Rate.}
%	\label{fig:CVER}
%\end{figure}

% YE: I shorten since we don't have space enough On top of keeping a good representation of the phonetic information in higher layers, the frozen SAMU-XLSR learned to keep semantic information until its last layer. The Figure \ref{fig:CER} shows that it only lose less than 1\% of CER with its final layer's generated embeddings. The CVER scores have the same behaviour.% (Figure \ref{fig:CVER}). 
We observe that the original frozen XLS-R model lost almost 7 points of CER between its best generated embeddings for semantic extraction task, layer 15, and its final generated embeddings, layer 24. 
%However, the XLS-R is originally a Wav2Vec 2.0 trying to minimize a contrastive loss and resolve a masking task. That is why it tends to update the middle-layer's embeddings to achieve its task, and lose some semantic information during the process through the final layer. 
On the other hand, since learning SAMU-XLSR consists on projecting its sentence-level embedding into the semantic multi-lingual LaBSE's encoding space, the highest layers of its encoder tend to capture and encode the semantics until the top layer.
%Therefore, it keeps all semantic information needed to be clustered together with text embeddings containing a great amount of semantic information. 
Both speech encoders give best CER results in middle layers. 
%\begin{figure}[htb]
%	\centering
%	\includegraphics[width=\columnwidth]{SAMUFR.png}
%	\caption{SAMU-XLSR Frozen layer-wise results}
%	\label{fig:SAMUFR}
%\end{figure}
%\begin{figure}[htb]
%	\centering
%	\includegraphics[width=\columnwidth]{SAMUFT.png}
%	\caption{SAMU-XLSR Fine-tuned layer-wise results}
%	\label{fig:SAMUFT}
%\end{figure}
%Figures \ref{fig:SAMUFR} and \ref{fig:SAMUFT} gives the precise evolution of all metrics for SAMU-XLSR, Frozen and Fine-tuned. 
As expected, fine-tuning the speech encoders allows the models to extract as much semantic information as possible from the audio signal. 
Even if the semantics extracted from the frozen SAMU-XLSR middle layers were already mostly kept through upper layers, performances were enhanced by fine-tuning the encoder.
%However, the WER scores indicate the loss of a lot of phonetic information in the upper layers of the frozen SAMU-XLSR. Fine-tuning the encoder greatly allows these information to be kept to the 24th layer. 
%\gl{HERE}

%\subsection{Injecting sentence-level embedding}

\subsection{Language portability}

\subsubsection{Zero shot}

To evaluate the multilingual portability of the SAMU-XLSR encoder compared to the original XLS-R, we apply zero-shot learning on our French (MEDIA) and Italian (PortMEDIA) data. 
We first train each end-to-end SLU model on the French data, by freezing or fine-tuning the speech encoder, and then make a simple inference on the Italian data. 
We also aim to measure how fine-tuning the speech encoder on the French data impacts the language portability capabilities.
%whether the fine-tuning on French data of the XLS-R module made in the SAMU-XLSR experiments helped portability through two languages of the same Latin family. 
\begin{table}[!ht]
    \begin{center}
    \newcolumntype{Y}{>{\centering\arraybackslash}X}
    \begin{tabularx}{\columnwidth}{| Y | Y | c | c | c | c |}
        \cline{3-6}    
        \multicolumn{2}{c |}{} & \textbf{ChER} & \textbf{WER} & \textbf{CER} & \textbf{CVER}\\
        \hline
        \multirow{2}{*}{\textbf{XLS-R}} & Frozen & 68.77 & 129.08 & 88.24 & 100.44\\
        & Fine-T. & 63.22 & 123.94 & 85.36 & 101.54\\
        \hline
        \multirow{1}{*}{\textbf{SAMU-}} & Frozen & 49.35 & 100.13 & 54.62 & 99.83\\
        \multirow{1}{*}{\textbf{XLSR}} & Fine-T. & 59.10 & 124.49 & 83.45 & 101.63\\
        \hline
        \end{tabularx}
        \caption{Zero-shot results (\%) from French MEDIA training to Italian PortMEDIA inference.}
        \label{tab:zeroX}
    \end{center}
\end{table}

Results in Table~\ref{tab:zeroX} show that the use of a frozen SAMU-XLSR speech encoder gives strongly better performance than other setups for concept recognition in these conditions: a CER of 54.62\% is attained while with the other configurations performs at more than 83\% error rate. 
We notice that, as expected, the performance related to the transcription itself is very bad: SAMU-XLSR is able to extract general semantics, but is not designed to provide language-dependent information useful to transcribe speech.
It also appears that fine-tuning SAMU-XLSR on French degrades the capability of the module to generate good semantic embeddings on Italian. 

%YE: I shorten since we need space
%It was not the case with the XLS-R speech encoder: a fine-tuning on same-family language enhanced the CER and other metrics of our Italian inference. 
%We can deduct that the SAMU-XLSR forced its embeddings from different languages to be represented in the same space clusters, thanks to LaBSE. 
%It allows the module to be a lot more efficient when dealing with multilingualism and language portability. 
\subsubsection{Low resource}

In these experiments, we exploit the training data in Italian to train the models.
IT in Table~\ref{tab:xPM} means the SLU model has been trained from scratch on the Italian data.
FR$\rightarrow$IT means the SLU model weights have been initialized with the French model before being trained on the Italian data.

%What we want to see here, is the language portability efficiency from French to Italian for the same SLU task by fine-tuning or freezing the same neural architecture.
%We processed by training our model with frozen and fine-tuned XLS-R and SAMU-XLSR with PortMEDIA to have a baseline. Then, we made the same experiments on MEDIA for 100 epochs, before continuing the training on PortMEDIA for 100 epochs. 

%At this point, we suspected the SAMU-XLSR to lose performances on ASR metrics for a specific language when fine-tuned on another one, even if they are from the same language family, because of the zero-shots results discussed previously.
%Indeed, the SAMU-XLSR can yet be considered efficient for Italian signal processing. Fine-tuning it on French could make it lose some efficiency regarding the suitability of its weights for the Italian transcription. 

\begin{table}[!ht]
    \begin{center}
    \newcolumntype{Y}{>{\centering\arraybackslash}X}
    \begin{tabularx}{\columnwidth}{| c | Y | c | c | c | c |}
        \cline{2-6}    
        \multicolumn{1}{c |}{} & \multirow{1}{*}{\textbf{Train}} & \multirow{2}{*}{\textbf{ChER}} & \multirow{2}{*}{\textbf{WER}} & \multirow{2}{*}{\textbf{CER}} & \multirow{2}{*}{\textbf{CVER}}\\
        \multicolumn{1}{c |}{} & \multirow{1}{*}{\textbf{Data}} & & & & \\
        \hline   
        \multirow{2}{*}{\textbf{Frozen}} & IT & 14.91 & 36.90 & 42.66 & 54.31\\ 
        %& FR$\rightarrow$IT & 17.14 & 46.06 & 42.50 & 57.81 \\ %lr 0.00001 
        & FR$\rightarrow$IT & 12.78 & 32.41 & 35.39 & 49.60 \\ 
        \hline
        \multirow{2}{*}{\textbf{Fine-T.}} & IT & 13.36 & 37.02 & 42.72 & 57.47 \\ 
        %& FR$\rightarrow$IT & 7.81 & 22.05 & 30.48 & 44.02 \\ %lr 0.00001
        & FR$\rightarrow$IT & 7.55 & 20.01 & 26.92 & 40.11 \\ 
        \hline
        \end{tabularx}
        \caption{XLS-R PortMEDIA results (\%) of PortMEDIA (IT) training and MEDIA training followed by PortMEDIA fine-tuning (FR$\rightarrow$IT).}
        \label{tab:xPM}
    \end{center}
\end{table}
Tables \ref{tab:xPM} and \ref{tab:sPM} illustrate the potential of portability of both encoders from French to Italian, with or without fine-tuning. 
We can observe that using SAMU-XLSR as a speech encoder still outperforms XLS-R, with a CER of 33.01\% (resp. 42.66\% for XLS-R) without fine-tuning and without the use of French data. 
With both fine-tuning and use of French data, SAMU-XLSR is able to reach 26.18\% of CER, but the gap with XLS-R -- that reaches 26.92\% -- is less significant.
\begin{table}[H]
    \begin{center}
    \newcolumntype{Y}{>{\centering\arraybackslash}X}
    \begin{tabularx}{\columnwidth}{| c | Y | c | c | c | c |}
        \cline{2-6}    
        \multicolumn{1}{c |}{} & \multirow{1}{*}{\textbf{Train}} & \multirow{2}{*}{\textbf{ChER}} & \multirow{2}{*}{\textbf{WER}} & \multirow{2}{*}{\textbf{CER}} & \multirow{2}{*}{\textbf{CVER}}\\
        \multicolumn{1}{c |}{} & \multirow{1}{*}{\textbf{Data}} & & & & \\
        \hline  
        \multirow{2}{*}{\textbf{Frozen}} & IT & 12.62 & 27.92 & 33.01 & 46.99 \\
        %& FR$\rightarrow$IT & 12.95 & 31.05 & 29.06 & 46.35 \\ %lr 0.00001
        & FR$\rightarrow$IT & 11.01 & 25.09 & 26.90 & 42.70 \\ 
        \hline
        \multirow{2}{*}{\textbf{Fine-T.}} & IT & 6.47 & 16.59 & 30.66 & 42.09 \\
        %& FR$\rightarrow$IT & 7.47 & 20.61 & 27.20 & 40.23 \\ %lr 0.00001
        & FR$\rightarrow$IT & 7.04 & 17.81 & 26.18 & 39.28 \\ 
        \hline
        \end{tabularx}
        \caption{SAMU-XLSR PortMEDIA results (\%) with PortMEDIA (IT) training and MEDIA training followed by PortMEDIA fine-tuning (FR$\rightarrow$IT).}
        \label{tab:sPM}
    \end{center}
\end{table}

%Fine-tuning the XLS-R model with only PortMEDIA results in less good Error Rates. This is because we are trying to improve the CER. When the ChER is our main metric, for an ASR task, the XLS-R is more suited for PortMEDIA data and fine-tuning it improve both CER and WER by ... \gl{need to get the results haha}. 
%Using MEDIA as a first fine-tune of our model, with a frozen encoder, does not improve the Error Rates too, except slightly for the CER. It shows that our system predicted French-Italian words because of an over-fitting on French data. It can not generate good Italian words, and mix them to French vocabulary. However, it is still able to recognise the semantic of the PortMEDIA data, even if it is not the same language. The CVER only went up due to the bad WER, but not because of the concepts recognition. 

%Considering the SAMU-XLSR, we can see that a fine-tuning on PortMEDIA improve all our different metrics. The task of Italian transcription being already managed by the initial encoder, our system does not have to improve it much and can focus on our main semantic recognition task. 
%We can see that doing a first training on French gives less good results in term of ASR metrics, as for the XLS-R before. The complete model is also over-fitting on French transcription.
%Nevertheless, we can see an important CER decrease when using MEDIA as first training, both for frozen and fine-tuned speech encoders. On top of that, frozen SAMU-XLSR allows the semantic transfer learning from French to Italian much more than its ancestor, the XLS-R. 

\subsection{Semantic analysis of sentence-level embeddings}
\label{sec:boc-semantic-analysis}

In the previous experiments, we analyzed the contents of the sequence embeddings produced by XLS-R and SAMU-XLSR. In this section, we examine if the pooled sentence embeddings produced by SAMU-XLSR contain enough semantic information according to the MEDIA and PortMEDIA tasks, and we analyze their cross-modal and cross-lingual abilities.
%Moreover, the SAMU-XLSR sentence embeddings are the representations that have been fine-tuned from XLS-R in order for them to be cross-modal with LaBSE text embeddings, and cross-lingual.
%By doing a semantic analysis on these sentence-levels embeddings, we should be able to see those cross-modal and cross-lingual abilities.
We simplify the MEDIA and PortMEDIA benchmark tasks to a bag-of-concepts classification task. 
For each segment, the system has to predict all the concepts that are present in the speech segment. We use a multi-hot representation for the output.
This simplification allows us to use the sentence embeddings without needing a complex decoder, that would bias our semantic analysis of the embeddings.

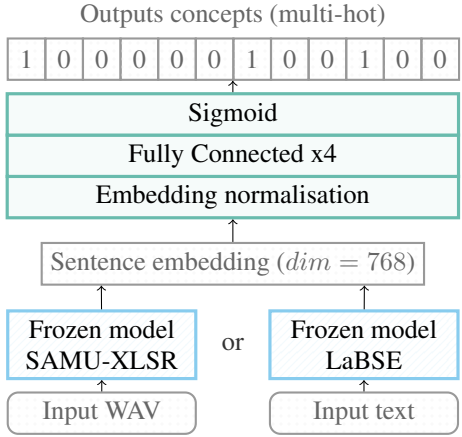
\begin{figure}[ht!]
    \centering
    \begin{tikzpicture}

    \node[invisible_node] (inputs) {};
    \node[io_node, right=0cm of inputs.west, minimum width=2.5cm] (inputs_wav) {Input WAV};
    \node[io_node, left=0cm of inputs.east, minimum width=2.5cm] (inputs_txt) {Input text};
    
    \node[model_node, above=0.15cm of inputs_wav, align=center, minimum width=2.5cm] (model_samu) {Frozen model\\ SAMU-XLSR};
    \node[model_node, above=0.15cm of inputs_txt, align=center, minimum width=2.5cm] (model_labse) {Frozen model\\ LaBSE};
    \node[invisible_node, minimum height=0.95cm, above=0.15cm of inputs] (model) {};
    \node[invisible_node, minimum height=0.95cm, above=0.15cm of inputs] (or) {or};

    \node[table, above=0.3cm of model] (pooled_embedding) {Sentence embedding ($dim = 768$)};
    \node[invisible_node, above right=0.3cm and 0cm of model.north west, minimum width=2.5cm] (pooled_embedding_left) {};
    \node[invisible_node, above left=0.3cm and 0cm of model.north east, minimum width=2.5cm] (pooled_embedding_right) {};

    \node[head_node, above=0.3cm of pooled_embedding] (embd_norm) {Embedding normalisation};
    \node[head_node, above=-\pgflinewidth of embd_norm] (fc) {Fully Connected x4};
    \node[head_node, above=-\pgflinewidth of fc] (sigmoid) {Sigmoid};

    \newcommand*\OutputsMultiHotList{{1, 0, 0, 0, 0, 0, 1, 0, 0, 1, 0, 0}}

    \foreach \x in {0,...,11}{
        \node[table_node, minimum height=0.55cm, above right=0.15cm and (\x * 0.5cm) - \figurewidth cm -\pgflinewidth of sigmoid] (output\x) {\pgfmathprint{\OutputsMultiHotList[\x]}};
    }
    \node[invisible_node, above=0.15cm of sigmoid] (output_all) {};

    % \node[loss, above=0.2cm of output_all] (outputs) {};
    \node[loss, above=0cm of output_all] (outputs) {Outputs concepts (multi-hot)};
    % \node[loss, above=0.1cm of outputs] (loss) {Binary Cross Entropy loss};

    \draw [->] (inputs_wav) edge (model_samu);
    \draw [->] (inputs_txt) edge (model_labse);
    \draw [->] (model_samu) edge (pooled_embedding_left);
    \draw [->] (model_labse) edge (pooled_embedding_right);
    \draw [->] (pooled_embedding) edge (embd_norm);
    \draw [->] (sigmoid) edge (output_all);
    
    \end{tikzpicture}
    \caption{Neural Architecture for an SLU language portability analysis of speech encoders with the MEDIA and PortMEDIA datasets.}
    \label{fig:architecture-sentence-level}
\end{figure}

Figure \ref{fig:architecture-sentence-level} illustrates the simple architecture we implemented for this sentence-level analysis. 
We extract the sentence embeddings of either SAMU-XLSR or LaBSE depending on the model we analyse. We apply $L_2$-normalisation on the embeddings as follows:
$\mathbf{x} \xleftarrow{} \frac{\sqrt{d}}{||\mathbf{x}||}\mathbf{x}$,
where $d=768$, i.e. the dimension of the embeddings. 
Fixing the norm (both in training and evaluation) is critical; unnormalized embeddings with large norm generate many false positives, while unnormalized embeddings with small norm generate many false negatives. Note also that the norm of the SAMU-XLSR embeddings may not be that informative, since the network is trained using cosine similarity between SAMU-XLSR and LaBSE embeddings, which is a norm-invariant objective function.
The network consists of four fully connected layers with ReLU activation functions and Dropout, and it is trained with weighted Binary Cross-Entropy loss.

In order to test both the cross-modal and cross-lingual properties of the embeddings, we train our classification models only on the French dataset. The Italian dataset is only used for testing, to obtain cross-lingual results.
For the cross-modal properties, we trained a model on SAMU-XLSR (speech) embeddings, and tested it on both SAMU-XLSR and LaBSE (text) embeddings. We also trained a second model on LaBSE embeddings in order to observe the difference when testing it with SAMU-XLSR speech embeddings.
The results, in terms of micro F$_1$-score are given in Table \ref{tab:sentence-level-results-all}. 
We did not test the XLS-R embeddings, as this model only provides frame-level embeddings.
We also report the frame-wise baseline results obtained in the previous experiments, by converting the sequence outputs of the models to a bag-of-concepts output. 
For these results, the model is trained to extract the sequence of concepts on the sequence of embeddings provided by SAMU-XLSR.

\begin{table}[ht!]
\centering
\newcolumntype{Y}{>{\centering\arraybackslash}X}
\begin{tabularx}{\columnwidth}{| c | Y | Y | Y |}
    \hline
    \multirow{2}{*}{\textbf{Test}} & \multirow{2}{*}{\textbf{Test}} & \multicolumn{2}{c |}{\textbf{Train}} \\
    \multirow{2}{*}{\textbf{Data}} & \multirow{2}{*}{\textbf{Encoder}} & \multicolumn{2}{c |}{\textbf{Encoder}} \\
    \cline{3-4}
    & & SAMU-XLSR & LaBSE \\
    \hline
    \multirow{3}{*}{FR} & SAMU-XLSR & 77.52\% & 71.77\% \\
    %\cline{2-4}
    & LaBSE & 78.04\% & 82.15\% \\ 
    \cdashline{2-4}
    & \textit{frame-wise}* & \textit{84.69\%} & - \\
    \hline
    \multirow{3}{*}{IT} & SAMU-XLSR & 68.55\% & 65.14\% \\
    %\cline{2-4}
    & LaBSE & 62.05\% & 69.58\% \\ 
    \cdashline{2-4}
    & \textit{frame-wise}* & \textit{59.76\%} & - \\
    \hline
\end{tabularx}
\caption{Micro F$_1$-scores for sentence-level semantic analysis with classification model trained on French and tested on French and Italian data. *Baseline results are obtained with the language portability models on frame-wise embeddings, and converted in bag-of-concepts outputs for evaluation.}
\label{tab:sentence-level-results-all}
\end{table}

We can observe that both LaBSE and SAMU-XLSR models obtain comparable results with their corresponding test embeddings when tested on the MEDIA dataset (77.52\% for SAMU-XLSR, 82.15\% for LaBSE).
The capacity of SAMU-XLSR to reproduce sentence-level embeddings close to the LaBSE ones is noticeable, and validates the strategy used to train SAMU-XLSR.

% However, SAMU-XLSR embeddings are more cross-modal and cross-lingual than LaBSE embeddings. The LaBSE classification head 
%Both the SAMU-XLSR and LaBSE classifiers are able to obtain good performances for the zero-shot cross-lingual PortMEDIA task (68.55\% and 69.58\% respectively).
 
 %For the zero-shot cross-modal scenario, we can see a small improvement in term of F$_1$ when training with SAMU-XLSR and testing with LaBSE embeddings, although this result is not reproduced on the never seen PortMEDIA data. We suggest that \valentin{in fact, I have no idea what could cause this}.

Finally, by comparing the results from the previous experiments with the frame-wise embeddings and this sentence-level embedding analysis, we observe that the sentence-level embeddings are better in extracting cross-lingual representations than the frame-level ones.
%This suggest that the alignment with the Language Agnostic embeddings (LaBSE) provided the SAMU-XLSR embeddings with good cross-lingual features that works well in this semantic analysis task. \valentin{I don't like this sentence, will need to find something else i think}
%\yk{I'm here}
\section{Conclusion}
%In this study, we investigated the use, for a complex SLU task, of a speech encoder built by combining the state-of-the-art multilingual acoustic frame-level XLS-R model with the Language Agnostic BERT Sentence Embedding model.
In this work, we investigated the capacity of the recently introduced SAMU-XLSR in addressing a challenging SLU task. SAMU-XLSR is a speech encoder is a fine-tuned version of the XLS-R model, using LaBSE embeddings as targets.   
In addition to its promising performance, we demonstrated how this speech encoder differs from the XLS-R model in the way it encodes the semantic information in its intermediate hidden representations. We also showed the real potential of the SAMU-XLSR for language portability.
Finally, we showed its capacity to build a sentence-level embedding able to highlight the semantic information of the task and its promising cross-lingual and cross-modal properties.

%\small

\section{Acknowledgements}
\vspace{-6pt}
\fontsize{8}{10.5}\selectfont
This work was performed using HPC resources from GENCI–IDRIS (grant 2022-AD011012565) and received funding from the European Union's Horizon 2020 research and innovation programme under the Marie Skłodowska-Curie ESPERANTO project (grant agreement N°101007666), through the SELMA project (grant N°957017) and from the French National Research Agency through the AISSPER project (ANR-19-CE23-0004).

\normalsize

\bibliographystyle{IEEEbib}
\small
\bibliography{strings,refs}

\end{document}